\def\year{2019}\relax

\documentclass[letterpaper]{article} 
\usepackage{aaai19}     
\usepackage{times}      
\usepackage{helvet}     
\usepackage{courier}    
\usepackage{url}        
\usepackage{graphicx}   
\usepackage{glossaries}
\usepackage{tabu}
\usepackage{booktabs}
\usepackage[table]{xcolor}
\usepackage{enumitem}
\usepackage{microtype}

\title{From FiLM to Video: Multi-turn Question Answering with Multi-modal Context}
\newcommand*\samethanks[1][\value{footnote}]{\footnotemark[#1]}
\author{
Dat Tien~Nguyen\textsuperscript{1}\thanks{Work performed during internship at Microsoft Research},~
Shikhar Sharma\textsuperscript{2}\thanks{These two authors contributed equally},~
Hannes Schulz\textsuperscript{2}\samethanks,~
Layla El~Asri\textsuperscript{2}\\
\textsuperscript{1} University of Amsterdam, Amsterdam, Netherlands \qquad
\textsuperscript{2} Microsoft Research, Montr\'eal, Canada\\
\url{t.d.nguyen@uva.nl}, \url{{shikhar.sharma,hannes.schulz,layla.elasri}@microsoft.com}
}

\newacronym{avsd}{AVSD}{Audio Visual Scene-aware Dialog}
\newacronym{amt}{AMT}{Amazon Mechanical Turk}
\newacronym{dstc7}{DSTC7}{Dialog System Technology Challenge 7}
\newacronym{lstm}{LSTM}{Long Short-Term Memory}
\newacronym{vqa}{VQA}{Visual Question Answering}
\newacronym{qa}{QA}{Question Answering}
\newacronym{visdial}{VisDial}{Visual Dialog}
\newacronym{modelname}{FA-HRED}{FiLM Attention Hierarchical Recurrent Encoder-Decoder}
\newacronym{hred}{HRED}{Hierarchical Recurrent Encoder-Decoder}
\newacronym{rl}{RL}{Reinforcement Learning}
\newacronym{igc}{IGC}{Image-Grounded Conversations}
\newacronym{cnn}{CNN}{Convolutional Neural Network}

\newcommand{\citet}[1]
{\citeauthor{#1}~\shortcite{#1}}
\newcommand{\citep}{\cite}

\newcommand\condprob[2]{\ensuremath{p\!\left(#1\,\middle|\,#2\right)}}

\begin{document}
\maketitle
\begin{abstract}
Understanding audio-visual content and the ability to have an informative conversation about it
have both been challenging areas for intelligent systems. The \gls{avsd} challenge, organized as
a track of the \gls{dstc7}, proposes a combined task, where a system has to answer questions
pertaining to a video given a dialogue with previous question-answer pairs and the video itself.
We propose for this task a hierarchical encoder-decoder model which computes a multi-modal
embedding of the dialogue context. It first embeds the dialogue history using two LSTMs. We
extract video and audio frames at regular intervals and compute semantic features using
pre-trained I3D and VGGish models, respectively. Before summarizing both modalities into
fixed-length vectors using LSTMs, we use FiLM blocks to condition them on the embeddings of the
current question, which allows us to reduce the dimensionality considerably. Finally, we use an
LSTM decoder that we train with scheduled sampling and evaluate using beam search. Compared to
the modality-fusing baseline model released by the \gls{avsd} challenge organizers, our model
achieves a relative improvements of more than 16\%, scoring 0.36~BLEU-4 and more than 33\%,
scoring 0.997~CIDEr.
\end{abstract}

\section{Introduction}
\label{sec:intro}
Deep neural networks have been successfully applied to several computer vision tasks such as image classification~\cite{ilsvrc2012}, object detection~\cite{objectDetection}, video action classification~\cite{actionClassification}, etc. They have also been successfully applied to natural language processing tasks such as machine translation~\cite{machineTranslation}, machine reading comprehension~\cite{mrc}, etc. There has also been an explosion of interest in tasks which combine multiple modalities such as audio, vision, and language together. Some popular multi-modal tasks combining these three modalities, and their differences are highlighted in Table~\ref{table:mm_tasks}.

Given an image and a question related to the image, the \gls{vqa} challenge~\cite{vqa} tasked users with selecting an answer to the question. \citet{vqaBias} identified several sources of bias in the \gls{vqa} dataset, which led to deep neural models answering several questions superficially. They found that in several instances, deep architectures exploited the statistics of the dataset to select answers ignoring the provided image. This prompted the release of \gls{vqa} 2.0~\cite{vqa2} which attempts to balance the original dataset. In it, each question is paired to two similar images which have different answers. Due to the complexity of \gls{vqa}, understanding the failures of deep neural architectures for this task has been a challenge. It is not easy to interpret whether the system failed in understanding the question or in understanding the image or in reasoning over it. The CLEVR dataset~\cite{johnson2017clevr} was hence proposed as a useful benchmark to evaluate such systems on the task of visual reasoning. Extending question answering over images to videos, \citet{tapaswi2016movieqa} have proposed MovieQA, where the task is to select the correct answer to a provided question given the movie clip on which it is based.
\begin{table}[h]
    \centering
    \begin{tabu}to\linewidth{@{}X[l,1.2]X[c]X[c]X[c,1.5]@{}} \toprule
        Task & Visual & Audio & Text Format \\ \cmidrule(r){1-1}\cmidrule(l){2-4}
        \gls{vqa} & Image & No & QA \\
        MovieQA & Video & Yes & QA \\
        VisDial & Image & No & QA-Dialogue \\  
        \gls{avsd} & Video & Yes & QA-Dialogue \\ \bottomrule
    \end{tabu}
    \caption{Tasks with audio, visual and text modalities}
    \label{table:mm_tasks}
\end{table}

Intelligent systems that can interact with human users for a useful purpose are highly valuable. To this end, there has been a recent push towards moving from single-turn \gls{qa} to multi-turn dialogue, which is a natural and intuitive setting for humans. Among multi-modal dialogue tasks, \gls{visdial}~\cite{visdial} provides an image and dialogue where each turn is a \gls{qa} pair. The task is to train a model to answer these questions within the dialogue. The \gls{avsd} challenge extends the \gls{visdial} task from images to the audio-visual domain.

We present our \gls{modelname} model for the \gls{avsd} task. \gls{modelname} combines a \gls{hred} for encoding and generating \gls{qa}-dialogue with a novel FiLM-based audio-visual feature extractor for videos and an auxiliary multi-task learning-based decoder for decoding a summary of the video. It outperforms the baseline results for the \gls{avsd} dataset~\cite{dstc7baseline} and was ranked $2^{nd}$ overall among the \gls{dstc7} \gls{avsd} challenge participants.

In Section~\ref{sec:related-work}, we discuss existing literature on end-to-end dialogue systems with a special focus on multi-modal dialogue systems. Section~\ref{sec:dataset} describes the \gls{avsd} dataset. In Section~\ref{sec:model}, we present the architecture of our \gls{modelname} model. We describe our evaluation and experimental setup in Section~\ref{sec:experiments} and then conclude in Section~\ref{sec:conclusion}.

\section{Related Work}
\label{sec:related-work}
With the availability of large conversational corpora from sources like Reddit and Twitter, there has been a lot of recent work on end-to-end modelling of dialogue for open domains. \citet{ritter} treated dialogue as a machine translation problem where they translate from the stimulus to the response. They observed this to be more challenging than machine translation tasks due the larger diversity of possible responses. Among approaches that just use the previous utterance to generate the current response, \citet{shang} proposed a response generation model based on the encoder decoder framework. \citet{sordoni} also proposed an encoder-decoder based neural network architecture that uses the previous two utterances to generate the current response. Among discriminative methods (i.e. methods that produce a score for utterances from a set and then rank them), \citet{lowe} proposed a neural architecture to select the best next response from a list of responses by measuring their similarity to the dialogue context. \citet{serbanHRED} extended prior work on encoder-decoder-based models to multi-turn conversations. They trained a hierarchical model called \gls{hred} for generating dialogue utterances where a recurrent neural network encoder encodes each utterance. A higher-level recurrent neural network maintains the dialogue state by further encoding the individual utterance encodings. This dialogue state is then decoded by another recurrent decoder to generate the response at that point in time. In followup work, \citet{serbanVHRED} used a latent stochastic variable to condition the generation process which aided their model in producing longer coherent outputs that better retain the context.

Datasets and tasks \cite{visdial,igc,guesswhat} have also been released recently to study visual-input based conversations. \citet{visdial} train several generative and discriminative deep neural models for the \gls{visdial} task. They observe that on this task, discriminative models outperform generative models and that models making better use of the dialogue history do better than models that do not use dialogue history at all. Unexpectedly, the performance between models that use the image features and models that do no use these features is not significantly different. As we discussed in Section~\ref{sec:intro}, this is similar to the issues ~\gls{vqa} models faced initially due to the imbalanced nature of the dataset, which leads us to believe that language is a strong prior on the \gls{visdial} dataset too. \citet{visdial_rl} train two separate agents to play a cooperative game where one agent has to answer the other agent's questions, which in turn has to predict the \textit{fc7} features of the Image obtained from VGGNet. Both agents are based on \gls{hred} models and they show that agents fine-tuned with \gls{rl} outperform agents trained solely with supervised learning. \citet{igc} train both generative and discriminative deep neural models on the \gls{igc} dataset, where the task is to generate questions and answers to carry on a meaningful conversation. \citet{guesswhat} train \gls{hred}-based models on GuessWhat?! dataset in which agents have to play a guessing game where one player has to find an object in the picture which the other player knows about and can answer questions about them.

Moving from image-based dialogue to video-based dialogue adds further complexity and challenges. Limited availability of such data is one of the challenges. Apart from the \gls{avsd} dataset, there does not exist a video dialogue dataset to the best of our knowledge and the \gls{avsd} data itself is fairly limited in size. Extracting relevant features from videos also contains the inherent complexity of extracting features from individual frames and additionally requires understanding their temporal interaction. The temporal nature of videos also makes it important to be able to focus on a varying-length subset of video frames as the action which is being asked about might be happening within them. There is also the need to encode the additional modality of audio which would be required for answering questions that rely on the audio track. With limited size of publicly available datasets based on the visual modality, learning useful features from high dimensional visual data has been a challenge even for the \gls{visdial} dataset, and we anticipate this to be an even more significant challenge on the \gls{avsd} dataset as it involves videos.

On the \gls{avsd} task, \citet{dstc7baseline} train an attention-based audio-visual scene-aware dialogue model which we use as the baseline model for this paper. They divide each video into multiple equal-duration segments and, from each of them, extract video features using an I3D~\cite{i3d} model, and audio features using a VGGish~\cite{vggish} model. The I3D model was pre-trained on Kinetics~\cite{kineticsDataset} dataset and the VGGish model was pre-trained on Audio Set~\cite{audioset}. The baseline encodes the current utterance's question with a \gls{lstm}~\cite{lstm} and uses the encoding to attend to the audio and video features from all the video segments and to fuse them together. The dialogue history is modelled with a hierarchical recurrent \gls{lstm} encoder where the input to the lower level encoder is a concatenation of question-answer pairs. The fused feature representation is concatenated with the question encoding and the dialogue history encoding and the resulting vector is used to decode the current answer using an \gls{lstm} decoder. Similar to the \gls{visdial} models, the performance difference between the best model that uses text and the best model that uses both text and video features is small. This indicates that the language is a stronger prior here and the baseline model is unable to make good use of the highly relevant video.
\begin{figure*}[t]
    \centering
    \includegraphics[width=.75\textwidth]{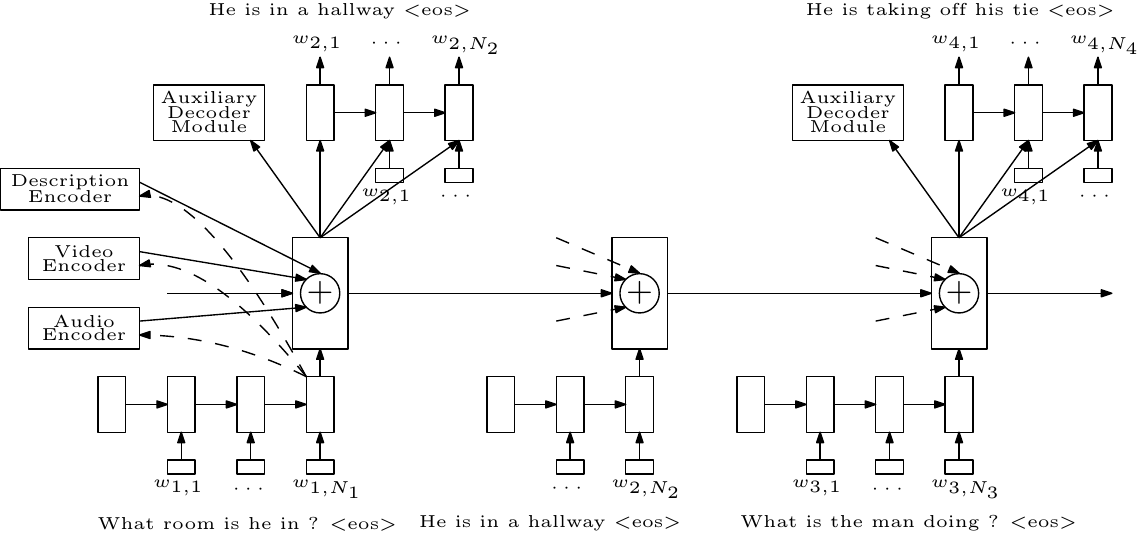}
	\caption{\gls{modelname} uses the last question's encoding to attend to video description, audio, and video features. These features along with the dialogue state enable the model to generate the answer to the current question. The ground truth answer is encoded into the dialogue history for the next turn.}
	\label{fig:model}
	\vspace*{-4mm}
\end{figure*}
\begin{figure}[t]
    \includegraphics[width=\columnwidth]{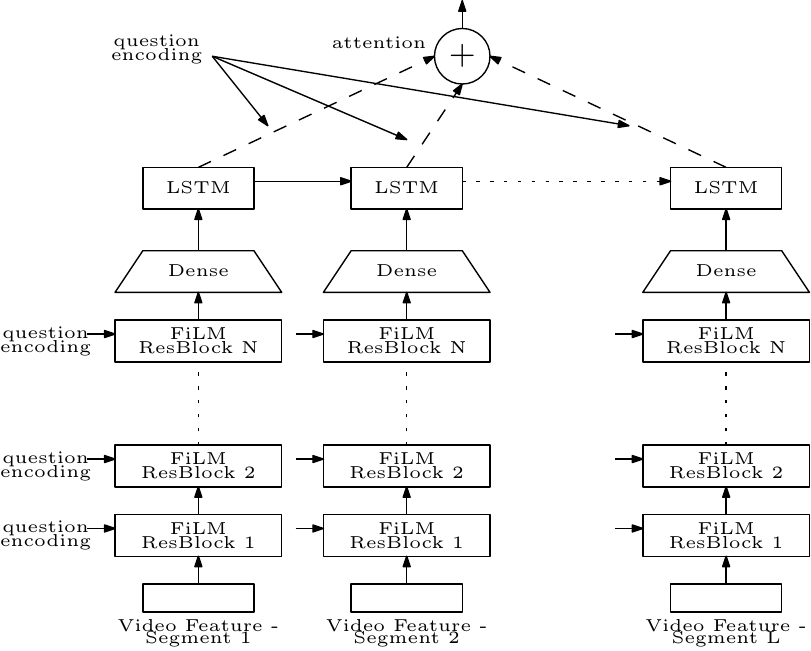}
    \caption{Video Encoder Module: FiLM for video features. Question encoding of the current question is used here.}
	\label{fig:film}
    \vspace*{-3mm}
\end{figure}
\begin{figure}[t]
    \includegraphics[width=\columnwidth]{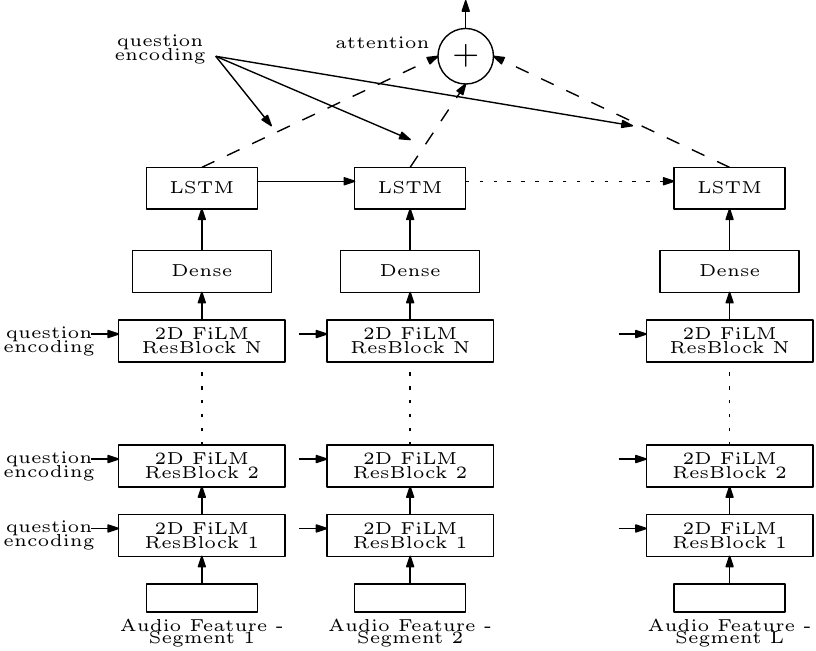}
	\caption{Audio Encoder Module: FiLM for audio features. Question encoding of the current question is used here.}
	\label{fig:film-audio}
	\vspace*{-3mm}
\end{figure}

Automated evaluation of both task-oriented and non-task-oriented dialogue systems has been a challenge \cite{taskeval,nontaskeval} too. Most such dialogue systems are evaluated using per-turn evaluation metrics since there is no suitable per-dialogue metric as conversations do not need to happen in a deterministic ordering of turns. These per-turn evaluation metrics are mostly word-overlap-based metrics such as BLEU, METEOR, ROUGE, and CIDEr, borrowed from the machine translation literature. Due to the diverse nature of possible responses, world-overlap metrics are not highly suitable for evaluating these tasks. Human evaluation of generated responses is considered the most reliable metric for such tasks but it is cost prohibitive and hence the dialogue system literature continues to rely widely on word-overlap-based metrics.

\section{The \gls{avsd} dataset and challenge}
The \gls{avsd} dataset~\cite{avsdDSTC7} consists of dialogues collected via \gls{amt}. Each dialogue is associated with a video from the Charades~\cite{charades} dataset and has conversations between two \gls{amt} workers related to the video. The Charades dataset has multi-action short videos and it provides text descriptions for these videos, which the \gls{avsd} challenge also distributes as the caption. The \gls{avsd} dataset has been collected using similar methodology as the \gls{visdial} dataset. In \gls{avsd}, each dialogue turn consists of a question and answer pair. One of the \gls{amt} workers assumes the role of \textit{questioner} while the other \gls{amt} worker assumes the role of \textit{answerer}. The \textit{questioner} sees three static frames from the video and has to ask questions. The \textit{answerer} sees the video and answers the questions asked by the \textit{questioner}. After 10 such \gls{qa} turns, the \textit{questioner} wraps up by writing a summary of the video based on the conversation.
\label{sec:dataset}

Dataset statistics such as the number of dialogues, turns, and words for the \gls{avsd} dataset are presented in Table~\ref{table:dataset}. For the initially released prototype dataset, the training set of the \gls{avsd} dataset corresponds to videos taken from the training set of the Charades dataset while the validation and test sets of the \gls{avsd} dataset correspond to videos taken from the validation set of the Charades dataset. For the official dataset, training, validation and test sets are drawn from the corresponding Charades sets.
\begin{table}[t]
    \centering
    \begin{tabu}to\linewidth{@{}X[l,1]*3{X[r]}@{}} \toprule
                     & training    & validation    & test      \\ \midrule
        \# dialogs   & 7\,659      & 1\,787        & 1\,710    \\
        \# turns     & 153\,180	   & 35\,740       & 13\,490   \\
        \# words     & 1\,450\,754 & 339\,006      & 110\,252  \\ \cmidrule(r){1-1}
        \# dialogs   & 6\,172      & 732           & 733       \\
        \# turns     & 123\,480    & 14\,680       & 14\,660   \\
        \# words     & 1\,163\,969 & 138\,314      & 138\,790  \\ \bottomrule
    \end{tabu}
    \caption{\gls{avsd}: Dataset Statistics. Top: official dataset. Bottom half: prototype dataset released earlier.}
    \label{table:dataset}
\end{table}

The Charades dataset also provides additional annotations for the videos such as action, scene, and object annotations, which are considered to be external data sources by the \gls{avsd} challenge, for which there is a special sub-task in the challenge. The action annotations also include the start and end time of the action in the video.

\section{Models}
\label{sec:model}
Our \gls{modelname} model is based on the \gls{hred} framework for modelling dialogue systems. In our model, an utterance-level recurrent \gls{lstm} encoder encodes utterances and a dialogue-level recurrent \gls{lstm} encoder encodes the final hidden states of the utterance-level encoders, thus maintaining the dialogue state and dialogue coherence. We use the final hidden states of the utterance-level encoders in the attention mechanism that is applied to the outputs of the description, video, and audio encoders. The attended features from these encoders are fused with the dialogue-level encoder's hidden states. An utterance-level decoder decodes the response for each such dialogue state following a question. We also add an auxiliary decoding module which is similar to the response decoder except that it tries to generate the caption and/or the summary of the video. We present our model in Figure~\ref{fig:model} and describe the individual components in detail below.

\subsection{Utterance-level Encoder}
The utterance-level encoder is a recurrent neural network consisting of a single layer of \gls{lstm} cells. The input to the \glspl{lstm} are word embeddings for each word in the utterance. The utterance is concatenated with a special symbol \texttt{<eos>} marking the end of the sequence. We initialize our word embeddings using 300-dimensional GloVe~\cite{glove} and then fine-tune them during training. For words not present in the GloVe vocabulary, we initialize their word embeddings from a random uniform distribution.

\subsection{Description Encoder}
Similar to the utterance-level encoder, the description encoder is also a single-layer \gls{lstm} recurrent neural network. Its word embeddings are also initialized with GloVe and then fine-tuned during training. For the description, we use the caption and/or the summary for the video provided with the dataset. The description encoder also has access to the last hidden state of the utterance-level encoder, which it uses to generate an attention map over the hidden states of its \gls{lstm}. The final output of this module is the attention-weighted sum of the \gls{lstm} hidden states.

\subsection{Video Encoder with Time-Extended FiLM}
For the video encoder, we use an I3D model pre-trained on the Kinetics dataset \citep{kineticsDataset} and extract the output of its \textit{Mixed\_7c} layer for $L$ ($30$ for our models) equi-distant segments of the video. Over these features, we add $N$ ($2$ for our models) FiLM~\cite{film} blocks which have been highly successful in visual reasoning problems. Each FiLM block applies a conditional (on the utterance encoding) feature-wise affine transformation on the features input to it, ultimately leading to the extraction of more relevant features. The FiLM blocks are followed by fully connected layers which are further encoded by a single layer recurrent \gls{lstm} network. The last hidden state of the utterance-level encoder then generates an attention map over the hidden states of its \gls{lstm}, which is multiplied by the hidden states to provide the output of this module. We also experimented with using convolutional \textit{Mixed\_5c} features to capture spatial information but on the limited \gls{avsd} dataset they did not yield any improvement. When not using the FiLM blocks, we use the final layer I3D features (provided by the \gls{avsd} organizers) and encode them with the \gls{lstm} directly, followed by the attention step. We present the video encoder in Figure~\ref{fig:film}.

\begin{table*}[t]
	\centering
	\begin{tabu}to\linewidth{@{}X[l,4.5]*7{X[c]}@{}} \toprule
		& BLEU-1 & BLEU-2 & BLEU-3 & BLEU-4 & METEOR & ROUGE & CIDEr \\ \midrule
		Challenge Baseline & 0.626 & 0.485 & 0.383 & 0.309 & 0.215 & 0.487 & 0.746 \\
		(Task 1.a.i) \gls{modelname} w/o summary & 0.648 & 0.505 & 0.399 & 0.323 & 0.231 & 0.510 & 0.843 \\
		(Task 1.a.ii) \gls{modelname} w/ summary & 0.695 & 0.553 & 0.444 & 0.360 & 0.249 & 0.544 & 0.997 \\
		(Task 2.a.i) \gls{modelname} w/o summary & 0.662 & 0.520 & 0.416 & 0.340 & 0.228 & 0.518 & 0.851 \\
		(Task 2.a.ii) \gls{modelname} w/ summary & 0.686 & 0.541 & 0.429 & 0.343 & 0.243 & 0.536 & 0.920 \\
		Winning Team (Task 1.a.ii) & {\bf 0.718} & {\bf 0.584} & {\bf 0.478} & {\bf 0.394} & {\bf 0.267} & {\bf 0.563} & {\bf 1.094} \\ \midrule
		
		(Task 1.a.i) \gls{modelname} w/o summary & 0.650 & 0.506 & 0.397 & 0.316 & 0.224 & 0.505 & 0.795 \\
		(Task 1.a.ii) \gls{modelname} w/ summary & {\bf 0.685} & {\bf 0.542} & {\bf 0.433} & {\bf 0.349} & {\bf 0.242} & {\bf 0.536} & {\bf 0.947} \\
		(Task 2.a.i) \gls{modelname} w/o summary & 0.635 & 0.500 & 0.398 & 0.323 & 0.220 & 0.501 & 0.799 \\
		(Task 2.a.ii) \gls{modelname} w/ summary & 0.656 & 0.507 & 0.398 & 0.319 & 0.228 & 0.513 & 0.836\\ \midrule
		
		Previous state-of-the-art~\cite{dstc7baseline}    & 0.256 & 0.161 & 0.109 & 0.078 & 0.113 & 0.277 & 0.727 \\
		(Task 1.a.i) \gls{modelname} w/o summary & 0.300 & 0.189 & 0.131 & 0.095 & 0.136 & {\bf 0.362} & 0.968 \\
		(Task 1.a.ii) \gls{modelname} w/ summary & {\bf 0.312} & {\bf 0.197} &	{\bf 0.136} & {\bf 0.098} &	{\bf 0.140} & {\bf 0.362} &	{\bf 0.993} \\
		(Task 2.a.i) \gls{modelname} w/o summary & 0.299 &	0.187 &	0.128 &	0.093 &	0.135 &	0.360 &	0.961 \\
		(Task 2.a.ii) \gls{modelname} w/ summary & 0.297 &	0.187 &	0.130 &	0.095 &	0.136 &	0.358 &	0.987 \\
		\bottomrule
	\end{tabu}
	\caption{Scores achieved by our model on different tasks of the \gls{avsd} challenge test set. Task 1 model configurations use both video and text features while Task 2 model configurations only use text features. \textbf{First section:} train on official, test on official. \textbf{Second section:} train on prototype, test on official. \textbf{Third section:} train on prototype, test on prototype.}
	\label{table:challenge_task_results}
\end{table*}

\subsection{Audio Encoder}
The audio encoder is structurally similar to the video encoder. We use the VGGish features provided by the \gls{avsd} challenge organizers. Also similar to the video encoder, when not using the FiLM blocks, we use the VGGish features and encode them with the \gls{lstm} directly, followed by the attention step. The audio encoder is depicted in Figure~\ref{fig:film-audio}.

\subsection{Fusing Modalities for Dialogue Context}
The outputs of the encoders for past utterances, descriptions, video, and audio together form the dialogue context $C$ which is the input of the decoder. 
We first combine past utterances using a dialogue-level encoder which is a single-layer \gls{lstm} recurrent neural network. The input to this encoder are the final hidden states of the utterance-level \glspl{lstm}.
To combine the hidden states of these diverse modalities, we found concatenation to perform better on the validation set than averaging or the Hadamard product.

\subsection{Decoders}
The \emph{answer decoder} consists of a single-layer recurrent \gls{lstm} network and generates the answer to the last question utterance. At each time-step, it is provided with the dialogue-level state and produces a softmax over a vector corresponding to vocabulary words and stops when 30 words were produced or an end of sentence token is encountered.

The \emph{auxiliary decoder} is functionally similar to the \textit{answer decoder}. The decoded sentence is the caption and/or description of the video. We use the Video Encoder state instead of the Dialogue-level Encoder state as input since with this module we want to learn a better video representation capable of decoding the description.

\subsection{Loss Function}
For a given context embedding $C_t$ at dialogue turn $t$, we minimize the negative log-likelihood of the answer word $0\leq w_{t,m}\leq V$ (vocabulary size), normalized by the number of words $M$ in the ground truth response $\mathbf r$,
\begin{align*}
    \mathcal L(C_t, \mathbf{r}) &\!=\! -\frac1M\sum_{m=1}^M\sum_i^V\left(
      [r_{t,m}{=}i]\log \condprob{r_{t,m}{=}i}{C_t, r_{t,m-1}^*}
    \right)
    ,
\end{align*}
where the probabilities $p(\cdot\,|\,\cdot)$ are given by the decoder LSTM output,
\begin{align*}
r^*_{t,m-1} &=\begin{cases}
    r_{t,m-1} & \hspace{-15mm};~s>0.2, s\sim U(0, 1)\\
    v \sim \condprob{r_{t,m-1}}{C_t,r^*_{t,m-2}} & \text{\hspace{5mm};~else}
\end{cases}
\end{align*}
is given by scheduled sampling \citep{scheduledSampling}, and $r^*_{t,0}$ is a symbol denoting the start of a sequence.
We optimize the model using the AMSGrad algorithm \citep{amsgrad} and use a per-condition random search to determine hyperparameters.
We train the model using the BLEU-4 score on the validation set as our stopping citerion.

\section{Experiments}
\label{sec:experiments}
The \gls{avsd} challenge tasks we address here are:
\begin{enumerate}[leftmargin=2\parindent]
    \item Video and Text
    \begin{enumerate}[label*=\alph*.,leftmargin=2\parindent]
        \item Allowed to use publicly available pre-trained models for feature extraction + \gls{qa} + captions
        \begin{enumerate}[label*=\roman*.,leftmargin=2\parindent]
            \item Not allowed to use summary
            \item Also allowed to use summary
        \end{enumerate}
    \end{enumerate}
    \item Text only
    \begin{enumerate}[label*=\alph*.,leftmargin=2\parindent]
        \item Allowed to use \gls{qa} + captions
        \begin{enumerate}[label*=\roman*.,leftmargin=2\parindent]
            \item Not allowed to use summary
            \item Also allowed to use summary
        \end{enumerate}
    \end{enumerate}
\end{enumerate}
We train our \gls{modelname} model for Task 1.a and Task 2.a of the challenge and we present the results in Table~\ref{table:challenge_task_results}. Our model outperforms the baseline model released by \citet{dstc7baseline} on all of these tasks. The scores for the winning team have been released to challenge participants and are also included. Their approach, however, is not public as of yet. We observe the following for our models:
\begin{itemize}
    \item \textbf{Comparing Task 1.a.i with 2.a.i and 1.a.ii with 2.a.ii:} Models that use both video and text data largely perform better than the models that just use the same text data. This is expected as the video features are highly relevant to answering the question asked. Based on findings presented by \citet{dstc7baseline} (which we discuss in Section~\ref{sec:related-work}), and due to the limited size of the dataset, we did not expect a large performance boost by adding the video features. As we can observe in Table~\ref{table:challenge_task_results}, the results are in line with these expectations.
    \item \textbf{Comparing Task 1.a.i with 1.a.ii and 2.a.i with 2.a.ii:} During the data collection phase, since the \textit{questioner} wrote the summary of the video at the end, we expected the summary to be extremely helpful in answering the \textit{questioner}'s questions. In line with this, we see that models that use the summary description encoder perform better than the models that use the caption description on almost all the metrics, including the BLEU-4 metric, the METEOR metric, and the CIDEr metric, all of which are highly popular metrics in the machine translation community. Any deviations from this trend have a very small margin. Evaluation metrics for such dialogue systems do not correlate much with human evaluation, as we discussed in Section~\ref{sec:related-work}, which makes this difference challenging to interpret. However, all of our models significantly improve upon the competitive baseline model.
\end{itemize}

\begin{table}[t]
\centering
\begin{tabu}to .95\linewidth{@{}X[l,4]X[c]X[c]@{}}
\toprule
    Model with (+) / without (-) inputs & \multicolumn1c{FiLM} & \multicolumn1c{No FiLM} \\ \midrule
    Attention + I3D + VGGish + Caption & 0.0986 & 0.0953\\
    \quad -Caption & 0.0951 & 0.0960\\
    \quad -Caption -VGGish & 0.0985 & 0.0974\\
    \quad -I3D -VGGish & -- & 0.0967\\
    \quad -I3D -Caption & 0.0960 & 0.0960\\
    \midrule
    Attention + I3D + VGGish + Summary & {\bf 0.1052} & 0.0989\\
    \quad -VGGish & 0.1045 & 0.1022\\
    \quad -I3D -VGGish & --  & 0.1007\\
    \quad -I3D & 0.1004 & 0.0999\\\bottomrule
\end{tabu}
\caption{Model ablation Study comparing BLEU-4 on the validation set: The best model makes use of all modalities and the video summary. Applying FiLM to audio and video features consistently outperforms unconditioned feature extraction. Video features (I3D) are more important than audio (VGGish). Combining all multi-modal components (e.g., text, audio and video) helps improve performance only when using FiLM blocks.}
\label{table:ablation}
\end{table}

Since the official test set has not been released publicly, results reported on the official test set have been provided by the challenge organizers. For the prototype test set and for the ablation study presented in Table~\ref{table:ablation}, we use the same code\footnote{\citet{taskeval}, {\scriptsize \url{https://github.com/Maluuba/nlg-eval}}} for evaluation metrics as used by \citet{dstc7baseline} for fairness and comparability. We attribute the significant performance gain of our model over the baseline to a combination of several factors as described below:
\begin{itemize}
    \item \textbf{Stronger encoder-decoder pair:} We use the \gls{hred} framework to compute utterance descriptors for questions and answers independently at the lower hierarchy level, instead of encoding their concatenation. This formulation removes the need for a separate question encoder and relies less on the long term memory of the utterance encoder. It improves validation performance by 0.015 BLEU-4 over a simple LSTM baseline.
    \item \textbf{Scheduled sampling:} In our experiments, we found that models trained with scheduled sampling performed better (about 0.004 BLEU-4 on validation set) than the ones trained using teacher-forcing for the \gls{avsd} dataset. Hence, we use scheduled sampling for all the results we report in this paper.
    \item \textbf{Regularization:} Due to the limited size of the \gls{avsd} dataset, we used dropout with retention probability between 0.6 and 0.9 for all our models during training to prevent over-fitting.
    \item \textbf{Feature extraction with FiLM:} We extract I3D and VGGish features using FiLM. The ablation study in Table~\ref{table:ablation} demonstrates that using FiLM provides a performance boost due to the extraction of features more relevant for the current question.
\end{itemize}

Our primary architectural differences over the baseline model are: not concatenating the question, answer pairs before encoding them, the auxiliary decoder module, and using the Time-Extended FiLM module for feature extraction. These, combined with using scheduled sampling and running hyperparameter optimization over the validation set to select hyperparameters, give us the observed performance boost.

We observe that our models generate fairly relevant responses to questions in the dialogues, and models with audio-visual inputs respond to audio-visual questions (e.g. ``is there any voices or music ?'') correctly more often.

We conduct an ablation study on the effectiveness of different components (eg.,
text, video and audio) and present it in Table~\ref{table:ablation}. Our experiments show that:
\begin{enumerate}
    \item Overall, models with audio, video, and summary performance better than models using audio, video, and caption.
    \item Using FiLM blocks for feature extraction consistently helps in improving performance across all settings. 
    \item Video features (I3D) are more important than audio (VGGish) for performance on the \gls{avsd} dataset.
    \item Caption and summary are the most important components. Interestingly, when not using FiLM, the performance decreases when combining text components with audio/video (0.1007 vs 0.0989 or 0.0967 vs 0.0953 ). By contrast, performance increases when using FiLM blocks to combine all multimodal components (0.1007 vs 0.1052 or 0.0967 vs 0.0986). We conclude that our FiLM implementation reduces overfitting compared to averaging or concatenation alternatives, when combining information from very different multiple modalities.
\end{enumerate}

\section{Conclusions}
\label{sec:conclusion}

We presented \gls{modelname}, a state-of-the-art dialogue model for conversations about videos. We evaluated the model on the official AVSD test set, where it achieves a relative improvement of more than 16\% over the baseline model on BLEU-4 and more than 33\% on CIDEr. The challenging aspect of multi-modal dialogue is fusing modalities with varying information density. On AVSD, it is easiest to learn from the input text, while video features remain largely opaque to the decoder. \gls{modelname} uses a generalization of FiLM to video that conditions video feature extraction on a question. However, similar to related work, absolute improvements of incorporating video features into dialogue are consistent but small. Thus, while our results indicate the suitability of our FiLM generalization, they also highlight that applications at the intersection between language and video are currently constrained by the quality of video features, and emphasizes the need for larger datasets.

\footnotesize
\bibliography{avsd}
\bibliographystyle{aaai}

\end{document}